\documentclass{article}

\usepackage{arxiv}

\usepackage[utf8]{inputenc} 
\usepackage[T1]{fontenc}    
\usepackage{hyperref}       
\usepackage{url}            
\usepackage{booktabs}       
\usepackage[usenames]{color}
\usepackage{amsfonts}       
\usepackage{nicefrac}       
\usepackage{microtype}      
\usepackage{graphicx}
\usepackage{natbib}
\usepackage{doi}
\usepackage{cite}
\usepackage{amsmath,amssymb,amsthm,amsthm}
\usepackage{algorithmic,algorithm,bm}
\usepackage{graphicx}
\usepackage{textcomp}
\usepackage{xcolor}
\usepackage{tikz}
\usepackage{pgfplots}
\usepackage{multicol}
\usepackage[usenames]{color}
\usepackage{float}
\usepackage{epsfig, multicol, rotating, multirow, pstricks, pst-grad}
\usepackage{pifont, relsize,  wrapfig, sidecap, bbding, pifont, array, comment}
\usepackage[justification=centering]{caption}

\pgfplotsset{compat=1.17}

\newtheorem{proposition}{Proposition}[section]

\newcommand{\param}{\theta} 
\newcommand{\standardwinlength}{L} 
\newcommand{\winlength}{N} 
\newcommand{\cols}{\mathbf{T}}
\newcommand{\complex}{j}

\newcommand{\idx}{b}
\newcommand{\col}{i}
\newcommand{\f}{f}
\newcommand{\loss}{\mathcal{L}}
\newcommand{\spec}{\mathcal{S}}

\definecolor{darkred}{rgb}{.82,0,0}
\definecolor{BleuBleu}{RGB}{0 50 150}
\definecolor{darkgreen}{rgb}{0,.4,0}

\title{A differentiable short-time Fourier transform\\ with respect to the window length}

\author{ 
 	Maxime Leiber{$^{*,\dagger}$}, Axel Barrau{$^\ddagger$}, Yosra Marnissi{$^\dagger$}, Dany Abboud{$^\dagger$} \vspace{.5cm} \\
 	{$^*$} INRIA, DI/ENS, PSL Research University \\
 	{$^\dagger$} Safran Tech, Groupe Safran \\
 	{$^\ddagger$} Offroad \\
}

\hypersetup{
pdftitle={A differentiable short-time Fourier transform with respect to the window length},
pdfsubject={STFT, neural networks, differentiable optimization, statistics},
pdfauthor={Maxime Leiber},
pdfkeywords={STFT, spectrogram, window length, backpropagation, differentiable optimization, neural network},
}

\begin{document}
\maketitle

\vspace{1cm}

\begin{abstract}
In this paper, we revisit the use of spectrograms in neural networks, by making the window length a continuous parameter optimizable by gradient descent instead of an empirically tuned integer-valued hyperparameter. The contribution is mainly theoretical at the moment, but plugging the modified STFT to any existing neural network is straightforward. 
We first define a differentiable version of the STFT in the case where local bins centers are fixed and independent of the window length parameter. We then discuss the more difficult case where the window length affects the position and number of bins. We illustrate the benefits of this new tool on an estimation and a classification problems, showing it can be of interest not only to neural networks but to any STFT-based signal processing algorithm.
\end{abstract}


\section{Introduction}
\label{sec:intro}
The short-time Fourier transform (STFT) is an important tool for analyzing non-stationary signals such as transient events like animal sounds \citep{Stafford}, discontinuous events like electroencephalography signals \citep{Ramos} as well as smoothly varying multi-harmonic signals typically measured on variable speed mechanical components \citep{Leclere}. Spectrograms built from the outputs of the STFT can be used for simple visualization and understanding of non-stationary signals. In this respect, several researchers have been working to improve the readability of spectrograms using some post-processing techniques such as reassignment \citep{Flandrin} and synchro-squeezing \citep{Thakur,Auger}. 

Spectrograms can be also combined with other processing methods to perform more advanced tasks. For instance, \citep{Yu} combines principal component analysis with vibration signals spectrograms for mechanical fault detection. Spectrograms have also been commonly used to feed transformers, recurrent and convolutional neural networks in various applications, e.g. identification and estimation \citep{Shea}, speech recognition \citep{Badshah,AST}, music detection \citep{Schluter}, electrocardiogram classification \citep{Huang}, data augmentation \citep{SA1}, source separation \citep{Defossez} etc. 
In this work, we are particularly interested in the latter applications i.e. those combining neural networks and spectrograms. In this framework, the window length of the spectrogram is usually a fixed parameter, empirically set by trial-and-error or to the default values of commonly used signal processing libraries, without thorough investigation or further justification. But this parameter deserves to be carefully set, as it determines the trade-off between the time and frequency resolution of the spectrogram, in accordance with Heisenberg's uncertainty principle. 

Our main contribution in this paper is a new paradigm for window length optimization we believe could become a standard way of tuning the window length of spectrograms used as input to neural networks. It consists in modifying the definition of the STFT operator to make the window length a continuous parameter w.r.t. which spectrogram values can be differentiated. The main idea is to break down the \textit{window length} parameter into two variables: a \textit{numerical window support} integer parameter and a \textit{time resolution} continuous parameter. This distinction has multiple consequences regarding the differentiability of the STFT and, in turn, gives the window length parameter a similar role to neural network weights.In fact, the differentiability proofs and backpropagation formulas provided in this paper allows a joint tuning of the neural network weights and spectrogram resolution (via its window size).

It is worth to note that the problem of searching the best window length is not new and several methods have been proposed in the literature such as \citep{Zhong,Kwok,Czerw}. However, we emphasize that our goal in this paper is not to propose the best way for optimizing the window size parameter, but instead a method for online optimization of the spectrogram resolution in a neural network with gradient algorithms. In that respect, the aforementioned methods are not adapted to our context as they are offline methods. To our knowledge, the only attempt to optimize the window length with a gradient descent is found in the recent work \citep{Zhao} using a Gaussian window and  which can be seen as a special case of the theory we propose. However, our paper goes further: the STFT transform is mathematically differentiable and all calculations with propagation and backpropagation formulas are provided. 

This paper is organized as follows. In Section \ref{ssec:definitions}, we start by giving some definitions and notations. In Section \ref{sec:diffspec} we introduce the modified STFT, being differentiable w.r.t. the window length. In Section \ref{sec:specbackprop} we extend the theory to the difficult case of STFT with fixed-overlap. In Section \ref{sec:experiments} we illustrate the effectiveness of our approach through two applications and we close our discussion in Section \ref{sec:ccl} with few concluding remarks.

\section{Definitions and notations}
\label{ssec:definitions}
All over this paper we will refer to the STFT as an operation taking a one-dimensional time series $s[t]$ as input and returning a 2-dimensional table $\spec[\col,\f]$, each column $\spec[\col,:]$ being the Discrete Fourier Transform (DFT) of a slice of length denoted by $\standardwinlength = 2^n$ of the signal $s$ going from an index $\idx_\col$ to an index $\idx_\col + \standardwinlength-1$, multiplied by a second sequence $h_\standardwinlength$ called ``tapering function''. This leads to the following formula we take as a definition of the STFT:
\begin{equation}
\label{eq::eq1}
\begin{aligned}
\spec[\col,f] & = \mathcal{F}(h_\standardwinlength s[\idx_\col:\idx_\col+\standardwinlength-1])[f] \\
 & = \sum_{k=0}^{\standardwinlength-1} h_\standardwinlength[k]s[\idx_\col+k] e^{-2\complex\pi kf/\standardwinlength},
 \end{aligned}
\end{equation}
where $\mathcal{F}(\cdot)$ denotes the Discrete Fourier Transform (DFT) operator, $\col$ is an integer expressing the index of the slice and $\idx_\col$ the associated starting point. Starting indices $b_i$ of time intervals are usually equally spaced, so we only have to set the first index $\idx_0$ and spacing $\Delta \idx$ between $\idx_i$ and $\idx_{i+1}$. This spacing is usually defined as a percentage of $\standardwinlength$, through a ratio $\alpha$ called \emph{overlapping}: $\Delta b = \lfloor \alpha \standardwinlength \rfloor$ where $\lfloor \cdot \rfloor$ denotes the integer part. Finally, several choices exist for the tapering function $h_\standardwinlength$. A common one is the \emph{Hann window} defined as:
\begin{equation}
\label{eq::eq2}
h_\standardwinlength[k] = \frac{1}{2} - \frac{1}{2}\cos( \frac{2 \pi k}{\standardwinlength-1} ).
\end{equation}

Having made notations and terminologies clear, we will move next to the core of this paper: building a STFT whose window length $\standardwinlength$ is a continuous parameter w.r.t. which spectrogram values are differentiable.

\section{Differentiable fixed-size STFT}
\label{sec:diffspec}
Bulding a differentiable version of STFT means writing a formula similar to Eq. \eqref{eq::eq1} where $\standardwinlength$ becomes a continuous parameter and $\spec[i,f]$ is differentiable w.r.t. $\standardwinlength$. We see that each term of the sum is already differentiable w.r.t. $\standardwinlength$ seen as a continuous parameter. If $b_i$ indices are assumed static the only obstacle is the presence of $\standardwinlength$ as a bound of the sum, which suggests breaking down $\standardwinlength$ into a \emph{numerical window support} $\winlength$ and a \emph{time resolution} $\param$:
\begin{equation}
\label{eq::eq3}
\spec[i,f] = \sum_{k=0}^{\winlength-1} h_{\winlength,\param}[k] s[b_i+k] e^{-2\complex\pi kf/\winlength}.
\end{equation}
where $h_{\winlength,\param}$ is a tapering function defined on $[0, N-1]$ but taking non-zero values in the interval $[\frac{\winlength-1-\param}{2} , \frac{\winlength-1+\param}{2}]$ where we have:
\begin{equation}
\label{eq::eq4}
h_{\winlength, \param}(k) = h_\param(k+\frac{\param-\winlength+1}{2})
\end{equation}

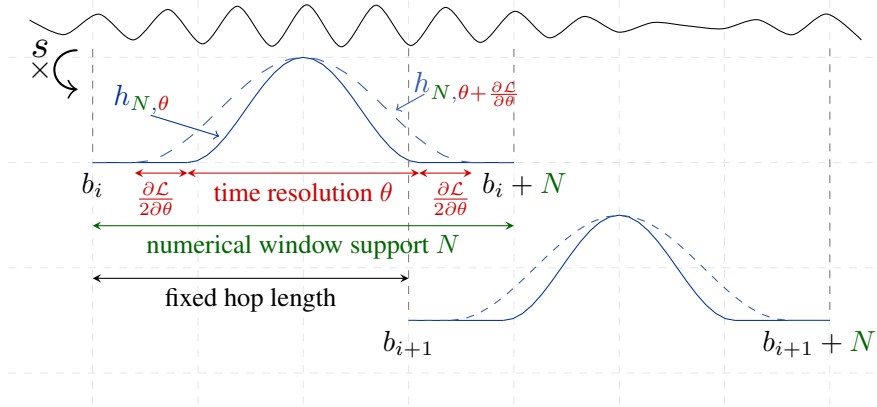
\begin{figure}[htbp]
\centering

\begin{tikzpicture}[scale=1.4]
\draw[help lines, very thin, color=gray!20, dashed] (-.8,-.3) grid (7.5,3.8); 
\draw[domain=-.6:7.3, smooth] plot (\x,{3.3+ .2*sin( 10 * \x r)}); 
\draw (-.5,3.3) node[below, scale=1.5] {$s$};
\newcommand*{\rotatecurvearrowleft}{\mathbin{\rotatebox{90}{$\curvearrowleft$}}}
\draw (-.25,2.85) node[scale=2] {$ \rotatecurvearrowleft $}; 
\draw (-.5,2.88) node[scale=1.2] {$\times$}; 

\draw[domain=.9:3.1, color=BleuBleu] plot (\x, { 2+  .5*(1-cos(2*pi*(\x-.9)/2.2 r))  });
\draw[domain=0:.9, color=BleuBleu] plot (\x, 2);
\draw[domain=3.1:4, color=BleuBleu] plot (\x, 2);
\draw[domain=.4:3.6, color=BleuBleu!80 , dash pattern=on 5pt off 5pt] plot (\x, {2+   .5*(1-cos(2*pi*(\x-.4)/3.2 r))  });
\draw[domain=0:.4, color=BleuBleu!80 , dash pattern=on 5pt off 5pt] plot (\x, 2);
\draw[domain=3.6:4, color=BleuBleu!80 , dash pattern=on 5pt off 5pt] plot (\x, 2);

\draw (0.1,2.6) node[right, scale=1.1] {\textcolor{BleuBleu}{$h_{\textcolor{darkgreen}{\winlength}, \textcolor{darkred}{\param}}$}};
\draw[-> , BleuBleu] (.55,2.45) -- (1.2,2.25);
\draw (2.95,2.7) node[right, scale=1.1] {\textcolor{BleuBleu!80}{$h_{\textcolor{darkgreen}{\winlength}, \textcolor{darkred}{\param + \frac{\partial \loss}{\partial \param} }}$}};
\draw[-> , BleuBleu!80] (3.15,2.65) -- (2.88,2.5);

\draw [dashed, gray](0,2.1) -- (0,3.1);
\draw [dashed, gray](4,2.1) -- (4,3.1);
\draw [dashed, gray](3,.6) -- (3,3.1);
\draw [dashed, gray](7,.6) -- (7,3.1);
\draw [darkred, stealth-stealth](0.9, 1.9) -- (3.1, 1.9);
\draw (2,1.9) node[below, darkred] {time resolution $\param$};
\draw [darkred, stealth-stealth](3.1, 1.9) -- (3.6, 1.9);
\draw [darkred] (3.4,1.9) node[below] {$\frac{\partial \loss}{2 \partial \param}$};
\draw [darkred, stealth-stealth](.4, 1.9) -- (.9, 1.9);
\draw [darkred] (.6,1.9) node[below] {$\frac{\partial \loss}{2 \partial \param} $};
\draw [stealth-stealth, darkgreen](0,1.4) -- (4,1.4);
\draw (2,1.4) node[below, darkgreen] {numerical window support $\winlength$};
\draw [stealth-stealth](0,.9) -- (3,.9);
\draw (1.5,.9) node[below] {fixed hop length};

\draw (0,2) node[below, scale=1.1] {$b_i$};
\draw (4.1,2) node[below, scale=1.1] {$b_i + \textcolor{darkgreen}{\winlength}$};
\draw (3,.5) node[below, scale=1.1] {$b_{i+1}$};
\draw (6.9,.5) node[below, scale=1.1] {$b_{i+1} + \textcolor{darkgreen}{\winlength}$};

\draw[domain=3.9:6.1, color=BleuBleu ] plot (\x, {.5+  .5*(1-cos(2*pi*(\x-3.9)/2.2 r))  });
\draw[domain=3:3.9, color=BleuBleu ] plot (\x, .5);
\draw[domain=6.1:7, color=BleuBleu ] plot (\x, .5);
\draw[domain=3.4:6.6, color=BleuBleu!80 , dash pattern=on 3pt off 3pt] plot (\x, {.5+   .5*(1-cos(2*pi*(\x-3.4)/3.2 r))  });
\draw[domain=3:3.4, color=BleuBleu!80 , dash pattern=on 3pt off 3pt] plot (\x, .5);
\draw[domain=6.6:7, color=BleuBleu!80 ] plot (\x, .5);

\end{tikzpicture}

\caption{Fixed-size STFT: on one hand, the size $\winlength$ of the subsignal on which DFT is computed is fixed, on the other hand, the support $\param$ of the tapering function, which actually determines time resolution is allowed to vary. Differentiation will be made w.r.t. this latter parameter.}
\end{figure}

Let us discuss a little bit the interpretation of $\winlength$ and $\param$. Parameter $\param$ has a meaning very close to the meaning of $\standardwinlength$: it sets the time length of the signal slice on which a local spectrum is computed. The difference with classical STFT is that the slice is filled with zeros in order to always be of size $\winlength$. As a consequence, frequency resolution of the local spectra is not anymore commended by $\param$ but by the second parameter $\winlength > \param$. This frequency resolution is unrealistic as the signal was padded with zeros: increasing $\winlength$ does not bring more information. Parameter $\winlength$ should only be seen as an upper bound on the time resolution $\param$ making differentiation possible. Now let us compute the differential of our proposed STFT w.r.t. the time resolution $\param$. We directly obtain:
\begin{equation}
\label{eq::eq5}
\frac{\partial \spec(\col,f)}{\partial \param} = \sum_{k=0}^{\winlength-1} \exp{ (-2\complex \pi \frac{kf}{\winlength})} \frac{\partial h(k)}{\partial \param} s[\idx_\col+k]
\end{equation}
where we recognize the STFT of $s$ with tapering function $h'= \frac{\partial h_{\winlength, \param}(k)}{\partial \param}$ instead of $h_{\winlength, \param}$:
\begin{equation}
\frac{\partial \spec[s]}{\partial \param} = \spec_{h'}[s]
\end{equation}
The latter result allows deriving compact gradient backpropagation formulas. We recall gradient backpropagation is the process of computing the derivative of a function $\loss$ w.r.t. the input of a function $g$ given its derivative w.r.t. the output of $g$, i.e., computing $\frac{\loss}{I}$ given $\frac{\loss}{\spec}$ for $ \spec = g(I)$. In our case we directly obtain:
\begin{align}
\label{eq::eq6}
\frac{\partial \loss}{\partial \param} &= \sum_{\col=1}^{\cols} \sum_{f=0}^{N-1} \frac{\partial \loss}{\partial \spec(\col,f)} \frac{\partial \spec(\col,f)}{\partial \param} \notag\\
&= \left \langle \frac{\partial \loss}{\partial \spec}, \spec_{h'}[s]  \right \rangle,
\end{align}
where $\langle \cdot , \cdot \rangle$ denoted the Froebenius scalar product of two matrices : $\lbrace A, B \rbrace = \sum_{i,j} A_{i,j} B_{i,j}$. 

Note we assumed until now that slices positions $[b_i, b_i+k]$ where independent from $\param$ meaning the number of columns of the STFT is constant. This case is useful when the STFT is the input to another algorithm that takes a fixed-size STFT as input, such as neural networks. But this is not representative of all use cases as the standard way to choose the indices $b_i$ is setting an overlapping ratio (usually 50\%) then letting indices $b_i$ depend on $\param$. This more elaborated case is discussed in Section \ref{sec:specbackprop} below.

\section{Differentiable fixed-overlap STFT}
\label{sec:specbackprop}
The more usual way of specifying the characteristics of a STFT is defining an overlapping parameter $\alpha$ instead of the bins positions. If window size is a fixed integer $L$ this is straightforward : the gap between starting indices of two slices is simply $\lfloor \alpha L \rfloor$. This is more complicated for differentiable STFT as setting this gap to $\lfloor \alpha \param \rfloor$ loses differentiability w.r.t. $\param$ while removing the integer part would mean the bound of the interval is not an integer. The solution is setting $b_i = \lfloor \alpha \param \rfloor$ and compensating the effect of the integer part by the combination of a small shift in the argument of $h_{\winlength,\param}$ and a factor $e^{2\complex\pi b_i f / \winlength}$ making the obtained STFT differentiable:
\begin{gather}
\label{eq::eq7}
\spec[i,f] = \sum_{k=0}^{\winlength-1} h_{\winlength,\param}(k - \text{frac}(i \alpha \param)) s[b_i+k] e^{\frac{-2\complex\pi (k+b_i)f}{\winlength}}
\end{gather}

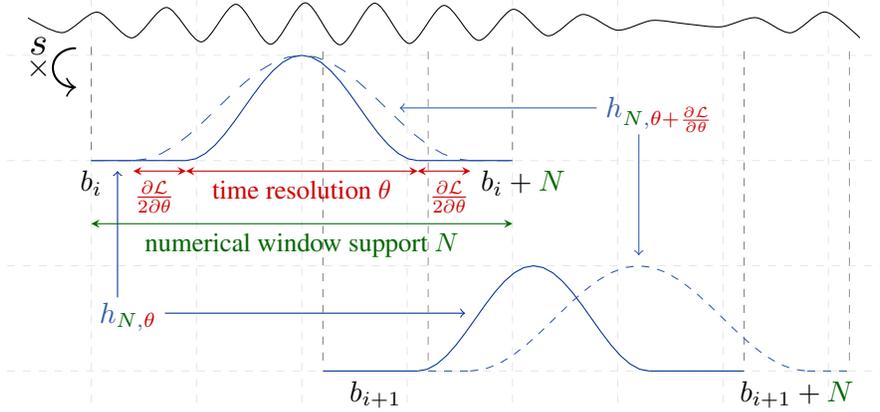
\begin{figure}[htbp]
\centering

\begin{tikzpicture}[scale=1.4]
\draw[help lines, very thin, color=gray!20, dashed] (-.8,-.3) grid (7.5,3.8); 
\draw[domain=-.6:7.3, smooth] plot (\x,{3.3+ .2*sin( 10 * \x r)}); 
\draw (-.5,3.3) node[below, scale=1.5] {$s$};
\newcommand*{\rotatecurvearrowleft}{\mathbin{\rotatebox{90}{$\curvearrowleft$}}}
\draw (-.25,2.85) node[scale=2] {$ \rotatecurvearrowleft $}; 
\draw (-.5,2.88) node[scale=1.2] {$\times$}; 

\draw[domain=.9:3.1, color=BleuBleu] plot (\x, { 2+  .5*(1-cos(2*pi*(\x-.9)/2.2 r))  });
\draw[domain=0:.9, color=BleuBleu] plot (\x, 2);
\draw[domain=3.1:4, color=BleuBleu] plot (\x, 2);
\draw[domain=.4:3.6, color=BleuBleu!80 , dash pattern=on 5pt off 5pt] plot (\x, {2+   .5*(1-cos(2*pi*(\x-.4)/3.2 r))  });
\draw[domain=0:.4, color=BleuBleu!80 , dash pattern=on 5pt off 5pt] plot (\x, 2);
\draw[domain=3.6:4, color=BleuBleu!80 , dash pattern=on 5pt off 5pt] plot (\x, 2);


\draw (.35,.75) node[below, scale=1.1] {\textcolor{BleuBleu!80}{$h_{\textcolor{darkgreen}{\winlength}, \textcolor{darkred}{\param}}$}};
\draw[-> , BleuBleu] (.25,.7) -- (0.25,1.9);
\draw[-> , BleuBleu] (.7,.55) -- (3.55,.55);
\draw (5.4,2.7) node[below, scale=1.1] {\textcolor{BleuBleu!80}{$h_{\textcolor{darkgreen}{\winlength}, \textcolor{darkred}{\param + \frac{\partial \loss}{\partial \param} }}$}};
\draw[-> , BleuBleu!80] (4.8,2.5) -- (2.93,2.5);
\draw[-> , BleuBleu!80] (5.2,2.25) -- (5.2,1.1);

\draw [dashed, gray](0,2.1) -- (0,3.1);
\draw [dashed, gray](4,2.1) -- (4,3.1);
\draw [dashed, gray](2.2,.1) -- (2.2,3.1);
\draw [dashed, gray!80](3.2,.1) -- (3.2,3.1);
\draw [dashed, gray](6.2,.1) -- (6.2,3.1);
\draw [dashed, gray!80](7.2,.1) -- (7.2,3.1);
\draw [darkred, stealth-stealth](0.9, 1.9) -- (3.1, 1.9);
\draw (2,1.9) node[below, darkred] {time resolution $\param$};
\draw [darkred, stealth-stealth](3.1, 1.9) -- (3.6, 1.9);
\draw [darkred] (3.4,1.9) node[below] {$\frac{\partial \loss}{2 \partial \param}$};
\draw [darkred, stealth-stealth](.4, 1.9) -- (.9, 1.9);
\draw [darkred] (.6,1.9) node[below] {$\frac{\partial \loss}{2 \partial \param} $};
\draw [stealth-stealth, darkgreen](0,1.4) -- (4,1.4);
\draw (2,1.4) node[below, darkgreen] {numerical window support $\winlength$};

\draw (0,2) node[below, scale=1.1] {$b_i$};
\draw (4.1,2) node[below, scale=1.1] {$b_i + \textcolor{darkgreen}{\winlength}$};
\draw (2.7,0) node[below, scale=1.1] {$b_{i+1}$};
\draw (6.7,0) node[below, scale=1.1] {$b_{i+1} + \textcolor{darkgreen}{\winlength}$};


\draw[domain=3.1:5.3, color=BleuBleu] plot (\x, { .5*(1-cos(2*pi*(\x-3.1)/2.2 r))  });
\draw[domain=2.2:3.1, color=BleuBleu] plot (\x, 0);
\draw[domain=5.3:6.2, color=BleuBleu] plot (\x, 0);
\draw[domain=3.6:6.8, color=BleuBleu!80, dash pattern=on 3pt off 3pt] plot (\x, {.5*(1-cos(2*pi*(\x-3.6)/3.2 r))  });
\draw[domain=3.2:3.6, color=BleuBleu!80, dash pattern=on 3pt off 3pt] plot (\x, 0);
\draw[domain=6.8:7.2, color=BleuBleu!80, dash pattern=on 3pt off 3pt] plot (\x, 0);

\end{tikzpicture}

\caption{Differentiable Fixed-overlap STFT: Fig. 2 has similar content to Fig. 1 but illustrates the difference with the fixed-size case: the center of the tapering function vary now with $\param$ and possibly akes a non integer value.}
\end{figure}

where $\text{frac} = x - \lfloor x \rfloor$. The remaining of this section will be dedicated to supporting this modified formula, first showing it is continuous w.r.t. $\param$, then showing it is even differentiable:

\begin{proposition}
STFT defined by Eq. (\ref{eq::eq7}) is everywhere continuous w.r.t. $\param$.
\end{proposition}
\begin{proof}
The integer and fractional part functions are continuous everywhere except on integers. So Eq. (\ref{eq::eq7}) is continuous everywhere except possibly when $i \alpha \param$ takes an integer value $p \in \mathbb{N}$. We will prove that it is also continuous then by looking at its left and right limits, i.e. for $\param = \frac{p}{i\alpha} \pm \epsilon$ with $  0$ going to zero.

For $\param = \frac{p}{i\alpha} - \epsilon$, we have $b_i = p-1$ and $\text{frac}(i \alpha \param)) = 1 - i \alpha \param$, which implies:
\begin{align}
\label{eq::eq8}
\spec[i,f] &= \sum_{k=0}^{\winlength-1} h_{\winlength,\param}(k-1+i \alpha \param) s[p-1+k] e^{\frac{-2\complex\pi (k+p-1)f}{\winlength}} \nonumber\\ 
 &= \sum_{k=0}^{\winlength-1} h_{\winlength,\param}(k'+ i \alpha \param)s[p+k']e^{\frac{-2\complex\pi (k'+p)f}{\winlength}}
\end{align}

To obtain the second line, we set $k'=k-1$, where $k'$ varies from $-1$ to $\winlength -2$. But first and last element being zero, the sum is not changed if $k'$ goes from $0$ to $\winlength -1$.

For $\param = \frac{p}{i\alpha} + \epsilon$, we have $b_i = p$ and $\text{frac}(i \alpha \param)) = i \alpha \param$, which implies:
\begin{align}
\label{eq::eq9}
\spec[i,f] = \sum_{k=0}^{\winlength-1} h_{\winlength,\param}[k-i \alpha \param] s[p+k] e^{\frac{-2\complex\pi (k+p)f}{\winlength}} 
\end{align}
So the limit is the same for $\param = \frac{p}{i \alpha} \pm \epsilon$ when $\epsilon \rightarrow 0$.
\end{proof}

We proved our differentiable fix-overlap STFT is continuous w.r.t. $\param$. Let us consider its differentiability. The first step is noticing that computations leading to Eq. (\ref{eq::eq6}) still hold, $h'$ being replaced by:
\begin{gather}
\label{eq::eq10}
h' = \frac{\partial}{\partial \param} \left[ h_{\winlength,\param}(k-\text{frac}(i \alpha \param)) e^{2\complex\pi b_i f / \winlength} \right]
\end{gather}
where $b_i = \lfloor i \alpha \param \rfloor$ is a function of $\param$. Eq. (\ref{eq::eq10}) is defined only when $i \alpha \theta$ is not an integer, i.e., when Eq. (\ref{eq::eq7}) is obviously differentiable. Extension to any value of $\param$ is the point of Prop. 3.2 below.

\begin{proposition}
The modified fixed-overlap STFT defined by Eq. (\ref{eq::eq7}) is everywhere differentiable w.r.t. $\param$.
\end{proposition}
\begin{proof}
If $i \alpha \param$ is not an integer, we have as in Eq. (\ref{eq::eq5}): 
\begin{equation}
\label{eq::eq11}
\frac{\partial \spec(\col,f)}{\partial \param} = \sum_{k=0}^{\winlength-1} \tilde h(k-\text{frac}(i \alpha \param)) {\partial \param} s[\idx_\col+k] e^{2\complex\pi (k+p) f / \winlength}
\end{equation}
with $\tilde h = - i \alpha \frac{\partial h_{\winlength,\param}}{\partial k} + \frac{\partial h_{\winlength,\param}}{\partial \param}$ obtain expanding Eq.(\ref{eq::eq10}). We recognize the same shape as Eq. (\ref{eq::eq7}) meaning a smimilar reasoning will show that the derivative Eq. (\ref{eq::eq11}) converges to the same limits when $i \alpha \theta$ goes to an integer by lower or upper values. As a consequence, the proposed fixed-overlap STFT is differentiable everywhere.
\end{proof}

We just proved that we actually obtain the desired differentiable w.r.t. the time resolution parameter $\theta$ version of STFT, even in the case of fix-overlap. This allows in particular gradient backpropagation, as Eq. (\ref{eq::eq6}) still applies. But fixed-overlap STFT raises a second issue: we assumed the number of columns of the obtained STFT was fixed, which is false as the number of columns increases when the time resolution decreases. The solution is padding the signal with zeros and taking into account all windows covering at least one value of the original signal. Then, the value of each column (replaced with zero when not defined) is a continuous function of $\param$. This way we obtained a STFT with fixed number of lines but a number of columns which depends on the chosen time resolution, akin to a classical STFT if overlapping ratio is maintained fix while window size changes.


\section{Experiments}
\label{sec:experiments}
In this section, we show how the proposed differentiable STFT can be of an immediate interest for any task involving a spectrogram. The first example illustrates the optimisation of the window length through a simple frequency tracking problem. In the second experiment, we investigate the joint optimisation of the window length with the weights of a simple neural network. 

The objective is to track the instantaneous frequency of a mono-component signal over time. The spectrogram is widely used in the literature to perform such a task. The motivation is that this difficult problem turns to be a peak detection and ridge tracking problem in the spectrogram: the aim is to detect the points of highest energy in the spectrogram plane. The challenge of time-frequency domain estimation is to find the appropriate window length to track fast frequency variations over time while maintaining a good frequency resolution. 

\begin{figure}[htbp]
    \centering
    \begin{minipage}[b]{.4\textwidth}
        \centering
        \centerline{\includegraphics[height=4.5cm]{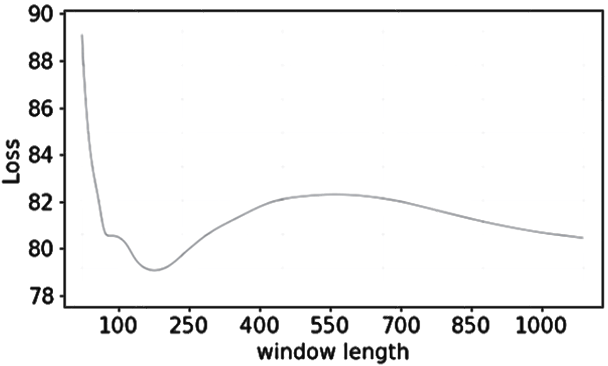}}
        \caption{Loss per window length. At the end of the gradient descent, the time resolution continuous parameter $\param$ reached the minimum value of the above cost function.}
        \label{fig:lossfreq}
    \end{minipage}%
    \hspace{1.5cm} \begin{minipage}[b]{.4\textwidth}
        \centering
        \centerline{\includegraphics[height=2.7cm]{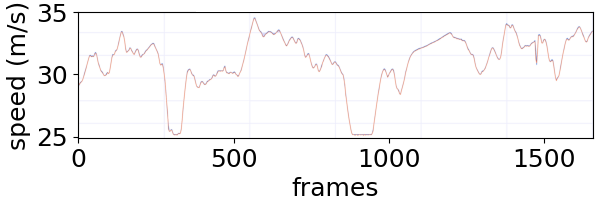}}
        \vspace{-.2cm} \caption{Spectrogram of a sample\\ obtained at the end of the training.}
        \label{fig:spec}
        
        \centerline{\includegraphics[height=2.7cm]{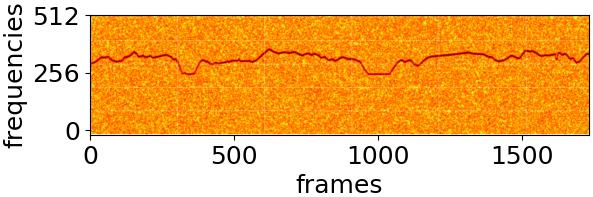}}
        \vspace{-.2cm} \caption{Spectrogram of a sample\\ obtained at the end of the training.}
        \label{fig:estimation}
    \end{minipage}
\end{figure}

In this example, we generate synthetic training data consisting of variable period sine plus white noise. The objective is to find the best window length that achieves the lowest mean square error in this training data
\begin{equation}
    \loss(\param) = \frac{1}{J}\sum_{j=1}^J\left\Vert\hat{y}_j^{\param}- \bar{y}_j\right\Vert^2
    \label{eq::15}
\end{equation}
where $\bar{y}_j$ is the true frequency of the $j^{th}$ signal of the training data and $\hat{y}_j^{\param}$ is the estimated frequency from the spectrogram  of window length $\param$ using some appropriate method. The latter can be a standard signal processing approach or a neural network including or not additional unknown parameters. For simplification reasons, we take in this example, the estimated frequency $\hat{y}[i]$ at each time interval $i$ as the weighted mean of the spectrogram frequencies with weights calculated from the spectrogram amplitude as follows \begin{equation}
    \hat{y}^{\param}[i]= \frac{\sum_f \spec^{\param}[i,f] f}{\sum_f \spec^{\param}[i,f]}
    \label{eq::14}
\end{equation}
where $\spec^{\param}$ is the spectrogram of window length $\param$. The latter is then the only unknown parameter of the loss function.

In our experiments, we consider our fixed size differentiable spectrogram and we run a gradient descend algorithm minimizing \eqref{eq::15} allowed by the backpropagation formulas of Sections \ref{sec:diffspec} and \ref{sec:specbackprop}. At the convergence of the gradient descent algorithm, the window length has reached the minimum of the loss on Fig. \ref{fig:lossfreq}. Fig. \ref{fig:spec} shows a spectrogram with the estimated window length and Fig. \ref{fig:estimation} the resulted estimated frequency. We see that the window length looks adapted to the problem although it was automatically tuned through gradient descent. This simple simulation proves the effectiveness of our backpropagation procedure based on a differentiable version of STFT. It also illustrates a general window length tuning methodology applicable to any existing signal processing algorithm involving spectrograms: replace the spectrogram computation step by the computation of our differentiable spectrogram, then optimise the window length by gradient descent based on the backpropagation formulas we have introduced.




\subsection{Joint optimisation with a neural network}
This second example illustrates how easy it is to plug our modified STFT into any existing neural network. Indeed, The goal of this experiment is to show that it is possible to jointly optimize the weights of a neural network with the window size of the spectrogram and that the latter will converge towards an optimal window size.

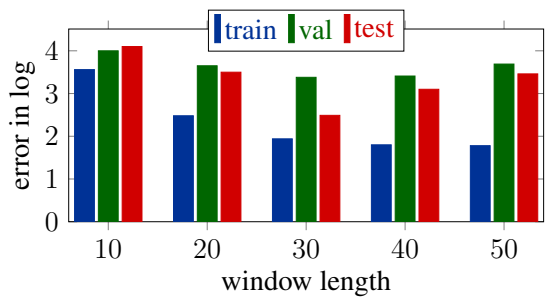
\begin{figure}[htbp]
\centering
\begin{tikzpicture}[scale=.75] \Large
\begin{axis}[
    ybar,
    height=5cm,
    width=10cm,
 	bar width = 10pt,
  	ylabel=error in log,
  	every axis y label/.style={at={(-0.05,.5)},rotate=90,anchor=near ticklabel},
  	xlabel=window length,
    every axis x label/.style={at={(.5,-.2)},anchor=near ticklabel},
    legend style={at={(.5,1.1)},rotate=90,anchor=north,legend columns=-1},
    legend image code/.code={\draw [#1] (0cm,-.2cm) rectangle (0.1cm,0.3cm); },
    ymin=0,
    xtick=data,
    ytick={0, 1, 2, 3, 4}
]
\addplot [draw=BleuBleu, fill=BleuBleu]
	coordinates {(10,3.56) (20,2.48)
		 (30,1.94) (40,1.80) (50,1.78)};
\addplot [draw=darkgreen, fill=darkgreen]
	coordinates {(10,4.) (20,3.65) 
		(30,3.38) (40,3.41) (50,3.69)};
\addplot [draw=darkred, fill=darkred]
	coordinates {(10,4.1) (20,3.5) 
		(30,2.49) (40,3.1) (50,3.46)};
		
\legend{\textcolor{BleuBleu}{train }, \textcolor{darkgreen}{val }, \textcolor{darkred}{test}}
\end{axis}
\end{tikzpicture}
 \caption{Training, validation and testing losses \\ of CNN per fixed window length.}
 \label{fig:table}
\end{figure}

\begin{figure}[htbp]
    \centering
    \begin{minipage}[b]{.4\textwidth}
        \centering
        \centerline{\includegraphics[height=5.6cm]{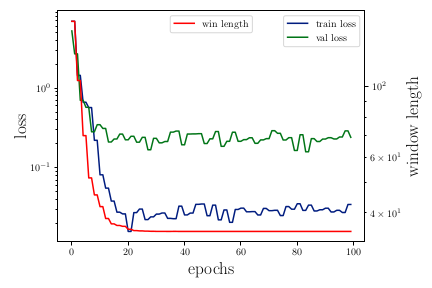}}
        \caption{Training loss, validation loss \\ and window length per epoch.}
        \label{fig:loss}
    \end{minipage}%
    \hspace{1.8cm} \begin{minipage}[b]{.4\textwidth}
        \centering
        \centerline{\includegraphics[height=6cm]{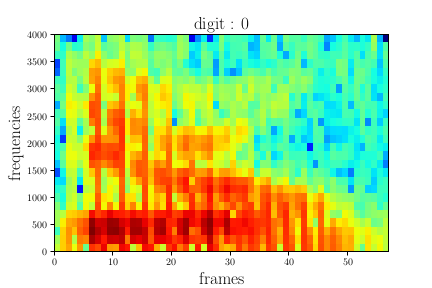}}
        \caption{Spectrogram of a sample\\ obtained at the end of the training.}
        \label{fig:windowlength}
    \end{minipage}
\end{figure}

We chose a simple classification task of spoken digit recognition using Free Spoken Digit Dataset (FSDD). 
The objective is to automatically find the best window size together with the best network weights for the considered dataset that minimizes the cross-entropy with the ground truth. We trained, for several previously fixed window sizes, a 2-layer convolutional neural network in order to compare the accuracy of the network as a function of the window size. We see in Fig. \ref{fig:table}, that the window size parameter is very important because the accuracy of the same trained network for different values of the window size varies strongly. Also that, whatever the initial value of the time resolution is, during a joint optimization with the network weights, the time resolution window length parameter converges to an optimal value as displayed on Fig. \ref{fig:loss}. Indeed, losses decrease by jointly optimizing the weights of the neural network and the time resolution continuous parameter $\param$ that converged to an optimal value: at the end of the gradient descent, the window length reached the value $34.9$ whereas we started the training with a window size of $200$. Fig \ref{fig:windowlength} shows the spectrogram of a sample at the end of the convergence.


\subsection{Discussion}

These two simple simulations prove the effectiveness of our backpropagation procedure based on a differentiable version of STFT. They illustrate a general window length tuning methodology applicable to any existing signal processing algorithm or neural network involving spectrograms: replace the spectrogram computation step by the computation of our differentiable spectrogram, then optimise the window length by gradient descent based on the backpropagation formulas we have introduced. The second example shows that is possible to jointly optimize a neural network with the window length of a spectrogram.

\section{Conclusion}
\label{sec:ccl}
We presented a modification of STFT making this operation differentiable w.r.t. the window length parameter, and gave the induced backpropagation formulas. As far as we know, this had not been done until now although this algorithm is ubiquitous in signal processing.
The main application of this contribution is revisiting the combination of spectrograms and neural networks: instead of directly giving a spectrogram as input to the network, one can now directly give the time signal as input to the network having our differentiable spectrogram as first layer and window length as continuous parameter and let it optimize this window length along with all weights of the network.

\bibliographystyle{unsrtnat}
\bibliography{references}  





\end{document}